# BCDDO: Binary Child Drawing Development Optimization

Abubakr S. Issa[1] , Yossra H. Ali[1], Tarik A. Rashid[2*]

[1]Computer Sciences Department, University of Technology, Baghdad, Iraq.
email:abubakr.s.issa@uotechnology.edu.iq; 110017@uotechnology.edu.iq
[2]Computer Science and Engineering, University of Kurdistan Hewler, Erbil, KR, Iraq.
Corresponding email: tarik.ahmed@ukh.edu.krd

**Abstract**

Child Drawing Development Optimization (CDDO) is a recently developed metaheuristic algorithm that has been demonstrated to perform well on multiple benchmark tests. In this paper, a Binary Child Drawing Development Optimization (BCDDO) is proposed for wrapper feature selection. The proposed BCDDO is utilized to choose a subset of important features to reach the highest classification accuracy. Harris Hawk optimization, Salp swarm algorithm, Grey Wolf optimization, and Whale optimization algorithm are utilized to evaluate the effectiveness and efficiency of the suggested feature selection method. In the field of feature selection to improve classification accuracy, the proposed method has gained a considerable classification accuracy advantage over previously mentioned methods. Four datasets are used in this research work; breast cancer, moderate covid, big covid, and Iris using XGboost classifier and the classification accuracies were (98.83%, 98.75%, 99.36%, and 96%) respectively for the Four mentioned datasets.

**Keywords:** Child Drawing Development Optimization, CDDO, BCDDO, Classification, Feature Selection.

## 1. Introduction

Data science has recently become a vital part of the healthcare industry. Typically, healthcare data are acquired from patients via electronic medical records. A typical application of data in healthcare is the development of decision support systems that incorporate patient data, domain knowledge, and artificial intelligence [1]. Even though ML models have been extensively examined and proved to be highly successful, disease prediction is a complex subject for which there are still numerous enhancements and approaches to investigate [1]. This type of problem falls within the categorization task of supervised learning within the machine learning field, which is relationship learning between a set of features and the target class using classification methods [2]. One of the significant goals of the classification of data is to predict based on the training data and available features. For machine learning tasks, large datasets with a high-dimensional feature space and a relatively limited sample size are crucial factors [3].

Dimensionality reduction is one of the main approaches for removing unnecessary and redundant features from the original feature set when there are a significant number of these features. Dimensionality reduction can enhance the effectiveness of a machine learning system and reduce its computing complexity by eliminating unnecessary and redundant information[4]. In prior years, feature selection and feature extraction were developed as solutions for dimensional reduction. Feature extraction creates fewer features by merging existing features so that these features include all (or the majority) of the information contained in the core features. Moreover, in feature selection, a subset of initial features is picked by deleting irrelevant and redundant features [6]. Filter schemes, wrapper schemes, and embedding schemes are the Four categories of feature selection classifiers. In the wrapper model, metaheuristic algorithms such as Bat Algorithm (BA), Particle Swarm Optimization (PSO), and SSA are used to evaluate a subset of features during their search operations to select a set of features with the maximum classification accuracy. The majority of wrapper approaches use iterative search procedures in which each iteration of the learning model guides the population of solutions toward the best solution to find the best feature subset [5].

The major contributions of this research work are:

1. Proposing a new binary version of the Child Drawing Development Optimization search algorithm to provide a way for efficient feature selection to increase resource usage, performance, and storage capacity while decreasing processing time.

2. Utilizing a modern classifier (XGboost) for classifying various types of datasets (Covid-19 images dataset, breast cancer numerical dataset).

3. Providing a firm framework for performance evaluation for the proposed method compared to the set of modern and competitor search algorithms (WOA, HHO, GWO, and SSA).

The arrangement of this paper's sections is as follows: Section 2 presents a summary of the works that are most relevant to our work. In Section 3, a brief overview was described. In section 4, the recommended methodology is provided. Section 5 establishes the results and the discussion. Comparative analysis with a few of the other methods is described in Section 6. In section 7, conclusions and further research are stated.

## 2. Related works

Feature selection is an important stage in the classification for the diagnosis and prediction of diseases. Several recent research in feature selection using search algorithms were reviewed in this research work. Yanan Zhang et al. [6] introduced a new version of HHO known as improved Harris Hawks optimization (IHHO) by combining the original Harris Hawks optimization with the Salp Swarm Algorithm to increase the optimizer's capacity to uncover high-quality global optimization and feature selection solutions. The results of IHHO compared to other feature selection approaches utilizing well-known benchmark datasets supplied by the University of California Irvine (UCI) indicate that the suggested IHHO produced higher accuracy rates than existing wrapper feature selection approaches, such as Binary Bat algorithm (BBA), Binary Salp Swarm algorithm (BSSA), and Binary Harris Hawk Optimization (BHHO) with the use of the KNN classifier. Kommana Swathi and Subrahmanyam Kodukula [7] suggested the Quantum Ant Lion (QAL) optimization to apply feature selection for gene classification and cancer diagnosis. The Quantum search process is used in the Ant Lion method to increase the search efficiency, which aids in increasing exploration and avoiding the local optima trap. In contrast, the Archimedes spiral search is used in the QAL method to increase exploitation in the feature selection based on the fitness function. Using the k-nearest neighbors algorithm (KNN), Support Vector Machines (SVM), Random Forests (RF), Deep Neural Networks (DNN), and Long-Short Term Memory (LSTM) classifiers on a gene dataset, 97.4% accuracy was achieved for gene classification.

Lingling Fang and Xiyue Liang [8] presented the Hyper Learning Binary Dragonfly Algorithm (HLBDA) as a wrapper-based technique for determining the best feature subset for different classification tasks. HLBDA is an enhanced version of the Binary Dragonfly Algorithm (BDA) that uses a hyperlearning technique to aid the algorithm in escaping local optima and improving its searching behavior. This article made use of 21 datasets from the Arizona State University and the University of California Irvine repository. Salima Ouadfel and Mohamed Abd Elaziz [9] suggested a high-dimensional feature selection technique utilizing the Relief filter method and a metaheuristic Equilibrium Optimizer (EO). The suggested approach is based on the Relief filter method and a Binary Equilibrium Optimizer (BEO) known as RBEO-LS, which consists of two parts. In the initial phase, the Relief technique is employed as a preprocessing step to give weights to features based on their estimated importance to the classification objective. BEO is utilized as a wrapper search strategy in the second phase. On sixteen UCI datasets and ten high-dimensional biological datasets, the performance of the created method has been assessed. The results also demonstrate the superiority of the RBEO-LS over other methods (Equilibrium Optimizer (EO), SSA, Sine Cosine Algorithm (SCA), Differential Evolution (DE), BAT algorithm, Binary particle swarm optimization (BPSO), Henry gas solubility optimization (HGSO)). Kushal Kanti Ghosh et al. [10] presented a feature selection technique based on a binary version of Manta Ray Foraging Optimization to address the problem of feature selection by employing eight transfer functions from two distinct families: S-shaped and V-shaped. The work employs 18 standard UCI datasets to test the performance and compare it to ALO-based methods and outperforms in terms of accuracy of the classification and features selected number. See details of the summary of the preview work in Table 1 below.

The research work is closely related to the work of Yanan Zhang et al. [8] and Lingling Fang and Xiyue Liang [10]. Like Zhang et al., this research also focuses on feature selection using a metaheuristic algorithm, but in this case, it is the Child. Child Drawing Development Optimization Furthermore, like Fang and Liang, a wrapper-based technique for feature selection is used, but a modified binary version of the CDDO algorithm is proposed that incorporates a new fitness function and a new update rule for the positions of the search agents. By comparing the results of the proposed method with other state-of-the-art feature selection techniques, it can be shown that the proposed method outperforms them in terms of accuracy and the number of selected features, which demonstrates the effectiveness of the proposed method in solving the feature selection problem.

*Table 1 Related Works Summary*

| Research work | Methods used | Limitations | Results |
|---|---|---|---|
| [6] | Improved Harris Hawks Optimization (IHHO) | 1-Randomness in Initialization: The randomness in the initialization process of candidate solutions can affect the adjusting capability of the Salp Chain introduced into HHO, potentially impacting optimization performance.<br><br>2-Computational Complexity: The computational complexity of IHHO may lead to increased optimization time, especially if the SSA strategy adjusts all individuals before each update. This could result in longer optimization times to attain high-quality solutions | The accuracy of the Improved Harris Hawks Optimization (IHHO) method in solving optimization problems was demonstrated through its superior performance compared to the original Harris Hawks Optimization (HHO) and other selected state-of-the-art algorithms on various benchmark functions |
| [7] | Quantum Ant Lion (QAL) optimization method | Sensitivity to parameter settings: The performance of the QAL method may be influenced by the selection of parameters, and finding optimal parameter values could be a challenging task.<br>Computational complexity: Implementing the QAL method may require significant computational resources, especially for large-scale gene datasets.<br>Interpretability: The complex nature of the QAL optimization process may make it difficult to interpret the results and understand the underlying feature selection mechanisms. | QAL method achieves an accuracy of 97.4% and outperforms the Ant Lion method, which has an accuracy of 86.2%. Additionally, when the selected features from QAL are applied to a Support Vector Machine (SVM) for classification, the QAL-SVM model shows higher efficiency in gene classification compared to other existing methods |
| [8] | Binary Dragonfly Algorithm | - | HLBDA in increasing prediction accuracy |

| | | | and reducing the number of selected features. |
|---|---|---|---|
| [9] | Enhanced Equilibrium Optimizer | - | In terms of accuracy, RBEO-LS had the best mean rank of 1.4688, outperforming other algorithms.<br>- For fitness value, RBEO-LS ranked first with a mean rank of 1.3438, followed by EO and BAT. |
| [10] | Manta ray foraging optimization | - | The results indicate that MRFO achieves high classification accuracy while selecting a lower number of features, making it a promising approach for feature selection in machine learning and data mining. |

## 3. Child Drawing Development Optimization algorithm

Child Drawing Development Optimization (CDDO) is a new metaheuristic algorithm The algorithm was designed by Sabat Abdulhameed and Tarik A. Rashid in 2021[11]. This algorithm is based on the child's learning behavior and cognitive development and uses the golden ratio to optimize the aesthetic value of their artwork. The golden ratio was first discovered by the renowned mathematician Fibonacci. The ratio of two consecutive numbers in the Fibonacci sequence is comparable; it is known as the golden ratio and is prevalent in nature, art, architecture, and design. CDDO simulates cognitive learning and the stages of a child's drawing development, beginning with the scribbling stage and proceeding to the pattern-based level of sophistication. Adjustments are made to the width, length, and golden ratio of the child's hand pressure to get better results. The following stages describe the Child Drawing Development Optimization algorithm.

A. The First Stage (The Scribble)
The initial attempts of a child to find drawing consist primarily of random marks. During this stage, the kid observes and discovers movement and hand pressure. Movement can be both linear and curved at random, as the child observes that liner movements result in lines and all other hand movements result in curves. In this step, the hand pressure is unreasonable; either too high or too low will be improved through trial and error in subsequent stages, taking into account a multitude of other factors—initialization $X_{ij}$ for I = 1 to N solutions. $X$ is the current solution reflecting a child's drawing with changeable choice variables such as hand pressure, golden ratio, length, and width of the drawing when multiple decision variables are considered. The number of decision variables is denoted by $I$ and the number of parameters is denoted by $j$.

B. The Second Stage (Exploitation)
During this stage, the kid learns to construct shapes by controlling the movement and direction of their bodies. One of the classifying factors for a child's performance is hand pressure. Initial Random Hand Pressure (*RHP*) is calculated using Equation (1). *RHP* is a random number between the lower boundary of the problem (LB) and the upper boundary of the solution (*UP*) used to compare the current solution's hand pressure with the current solution's hand pressure (*HP*). When *HP* is hand pressure and *j* is a collection of solution parameters, *HP* will be selected from the solution parameters using equation (2).

$$RHP = rand(LB, UP) \tag{1}$$

$$HP = X(i, rand(j)) \tag{2}$$

C. Third Stage (Golden Ratio)

The child is now at the stage where he or she applies the abilities acquired through experience and uses the feedback to study the pattern in the real pictures, attempt to assign meanings to the drawings, and practice drawing by copying, practicing, and being enthusiastic (with trail). The current hand pressure is compared to *RHP*; if it is less than *RHP*, the solution is updated using Eq. (2), taking into consideration the kid's skill rate (*SR*) and level rate (*LR*), which are two random integers between 0 and 1 initially and between 0.6 and 1 if the child has relevant hand pressure. Setting *SR* and *LR* to high values (0.6–1) suggests that the child's level of knowledge and competence is correct, however, it can be improved by factoring in the *GR* component. The Golden Ratio is another component utilized to update and enhance the performance of the solution (*GR*). *GR* is the ratio of the two selected solution components, which are the length and breadth of a child's artwork (see Eq. (3)). Each of these two components is picked at random from all of the problem's factors using (Eq. (4)).

$$X_{iGR} = \frac{X_{iL} + X_{iW}}{X_{iL}} \tag{3}$$

$$W, L = rand(0, j) \tag{4}$$

In Eq. (5), $X_{ilbest}$ represents the child's best drawing to date, which is the local finest solution, and $X_{igbest}$ represents the global best solution observed by the children in their surroundings, in addition, the Golden Ratio (*GR*) is the ratio between the child's drawing's length (*L*) and width (*W*).

$$X_{i+1} = GR + SR.* (X_{ilbest} - X_i) + LR.* (X_{igbest} - X_i) \tag{5}$$

D. Fourth Stage (Creativity)

During this phase, the child combines information to update golden-ratio or nearly golden-ratio solutions. However, the solution lacks meaningful hand pressure, indicating that a child's talents are underdeveloped and require further development through the application of the creative factor and the golden ratio. Pattern Memory (*PM*) is created for every solution in the algorithm; the size of the pattern varies based on the problems. Nonetheless, selecting a random solution from the *PM* array to be used for updating the solutions that are not performing well is one of the strategies to increase the convergence rate of the algorithm, and in practice, it speeds up the children's learning rate. In Eq. (6), both *CR* and *PM* are used to update the existing solution and converge on the optimal solution.

$$X_{i+1} = X_{iMP} + CR.* (X_{igbest}) \tag{6}$$

E. Fifth Stage (Pattern Memory)

This stage focuses primarily on drawing inches of detail. As the behavior is exhibited by the agent's best-updating mechanism, it is incorporated into the algorithm. It is when the answer will be modified if a better one becomes available, and the same is true for updating the population's best global solution. This will also be the case when refreshing the pattern memory with the best global solution attained thus far in each iteration. Figure 1 illustrates the steps of the CDDO algorithm.

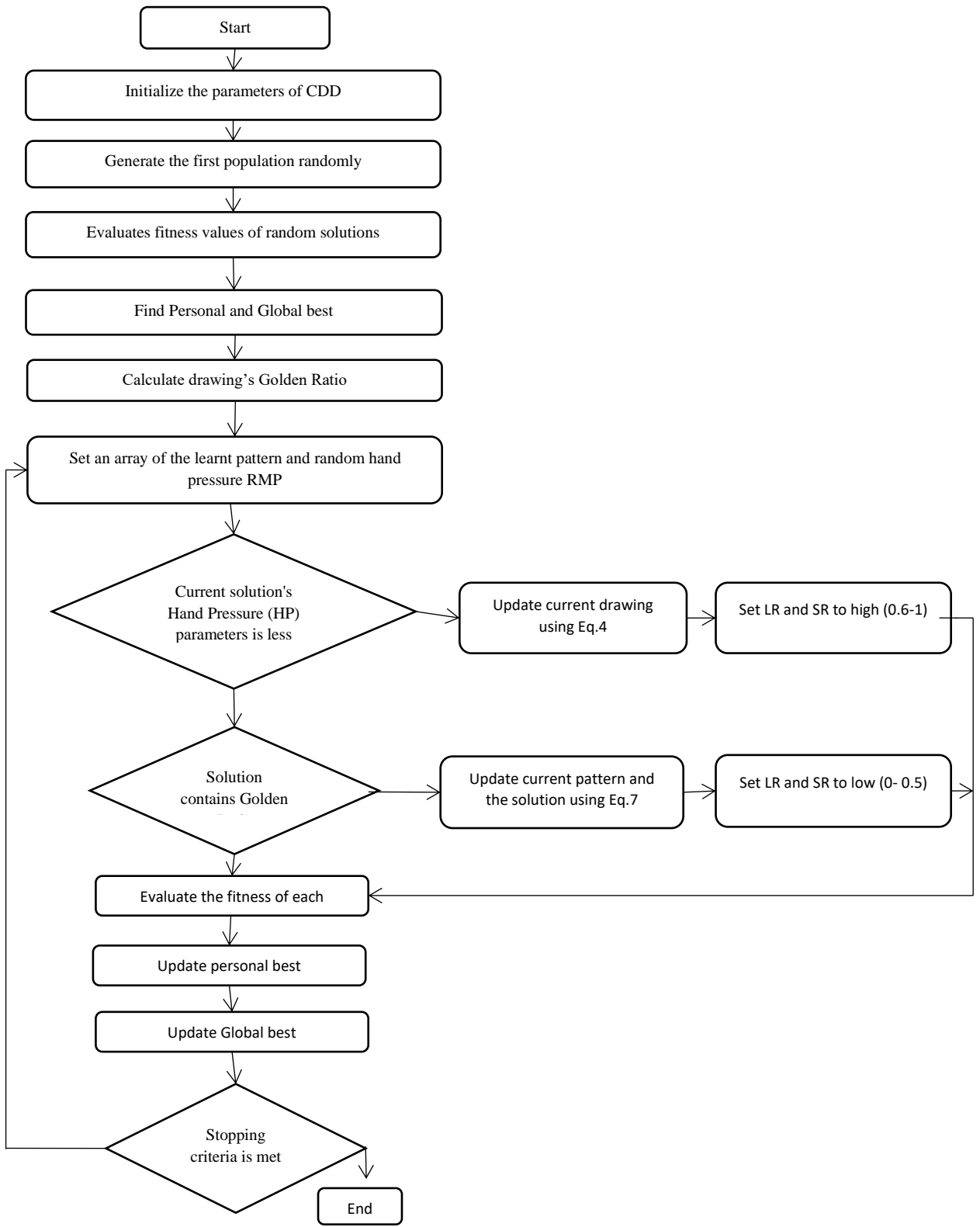

*Figure 1: flowchart of CDDO algorithm*

| |
|---|
| *Pseudocode of Child Drawing Development Optimization [11]* |
| *Begin* |
| *Initialize child's drawing population Xi (i = 1, 2, … , j)* |
| *Compute each drawing's fitness, set personal best and global best* |
| *Calculate the golden ratio of each drawing eq. (5)* |
| *Create a pattern memory array* |
| *Randomly choose an index of pattern memory* |
| *While (t < maximum number of iterations)* |
| *Calculate RHP using eq. (2)* |
| *Randomly choose hand pressure P1 eq. (3) Length P2, Width P3 eq. (6)* |
|     *For each drawing* |
|         *if (hand pressure was low)* |
|         *Update the drawings using eq. (4)* |
|         *Set LR and SR to HIGH (0.6-1)* |
|         *End if* |
|         *Elseif (XiGR is near to golden ratio)* |
|         *Consider the learned patterns, LR and SR using eq. (7)* |
|         *Set LR and SR to LOW (0-0.5)* |
|         *Endif* |
|         *Evaluate the cost values* |
|         *Update Personal, global best* |
|         *Update pattern memory* |
|         *Store the Best Cost Value* |
| *End for* |
| *Increment t* |
| *End while* |
| *End* |
| *Return Global best* |

**4. Binary Child Drawing Development Optimization**

In the continuous Child Drawing Development Optimization, the values of drawings continually transfer to a different location in the search space. In some unique situations, such as feature selection, the solutions are limited to the binary 0 or 1 value, motivating a special version of the CDDO algorithm. In this paper, a novel binary Child Drawing Development Optimization for feature selection is proposed.4.1 introduces the representation used in the binary Child Drawing Development Optimization algorithm for feature selection. It explains the binary discrete search space that is inherent to feature selection problems and how a vector of 0's and 1's is used to represent solutions to such problems. While 4.2 discusses the fitness function used in the BCDDO algorithm for feature selection. It explains how the evaluation of the selected feature subset is a critical element of feature selection methods and involves a classifier, which in this case is the K-Nearest Neighbor classifier, The KNN is chosen due to its promising performance and fast computation speed in previous work [12].

### 4.1 BCDDO: Representation

A feature selection issue has its own inherent binary discrete search space. To express any solution to a feature selection problem, a vector of 0's and 1's is required, where 0 indicates the relevant feature not selected and 1 represents selected [13]. The CDDO is utilized to determine the most informative subset of features. The CDDO initiates the feature selection by generating ($N \times D$) initial solutions, where $N$ is the number of particles and $D$ is the number of features. Each vector in the population reflects the indices of associated traits. A threshold of 0.5 is used to decide whether or not the feature is selected.

$$\begin{cases} x_i^d > 0.5, & \text{Feature will be elected} \\ x_i^d \leq 0.5, & \text{Feature will be deprecated} \end{cases} \quad (7)$$

According to Eq. (7), if the vector's value is larger than 0.5, the related feature is selected. Otherwise, the feature is not chosen.

Figure 2 presents a binary representation of a possible BCDDO solution for a data set with six characteristics, of which four are chosen and the other two are not.

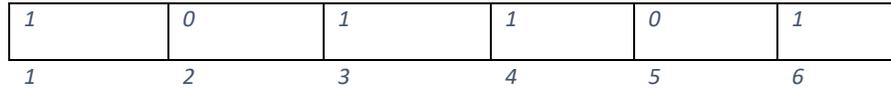

*Figure 2: Binary illustration of a possible BCDDO*

### 4.2 BCDDO: Fitness Function

Evaluation of the selected feature subset is a critical element of feature selection methods. The evaluation methodology involves a classifier because the suggested method is a wrapper-based feature selection technique. For this objective, we employed the well-known K-Nearest Neighbor classifier. Generally, FS has two goals: greater classification accuracy and a smaller number of selected features. The superiority of the selected subset is indicated by both the greater classification accuracy and the reduced number of selected features. During the development of the fitness function for the proposed FS method, we examined both aspects. We chose classification error, the complement of classification accuracy since minimizing the number of features is important. Eq. 8 defines the fitness function used

$$fitness = a * cls_{err} + b * (\frac{feature\ selected}{max\ number\ of\ features}) \quad (8)$$

Where a = 0.90 is a constant for managing accuracy, *b* random integer improves accuracy calculated using Eq. 9, and $cls_{err}$ is the classification error rate of KNN. Figure 2 illustrates the block diagram of the proposed BCDDO classification system, and Algorithm 1 describes the steps of the binary version of CDDO (BCDDO)

$$b = 1-a \qquad (9)$$

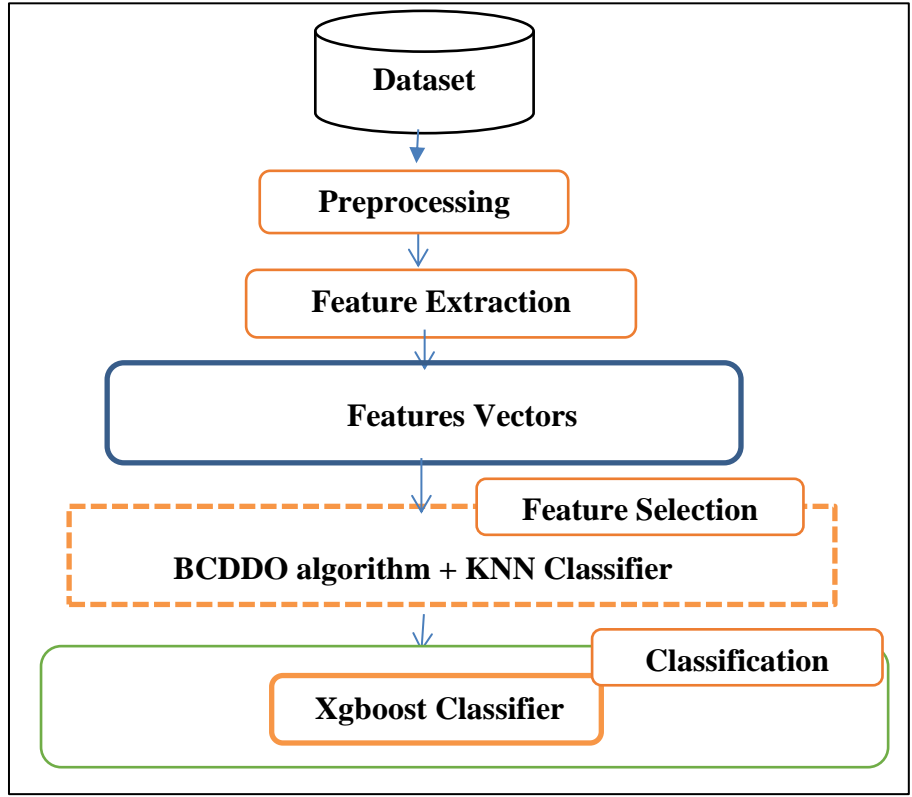

Figure 3 Block diagram of the proposed BCDDO classification system

| Algorithm1: Binary Child Drawing Development Optimization |
|---|
| Input: Set of features |
| Output : Subset of best features |
| Begin<br>    Set the initial drawing population of the child $X_i$ $(i = 1, 2, ..., j)$<br>    Represent each drawing as the feature vector<br>    Convert each feature vector to a binary vector using eq. (7)<br>    Calculate the fitness of each vector using eq. (8)<br>    Establish personal and global best<br>    Determine the golden ratio of each drawing utilizing the eq. (3)<br>    Initialize an array of pattern memory<br>    Select a pattern memory index randomly<br>    While ($t$ < maximum number of iterations)<br>        Compute *RHP* value by using eq. (1)<br>        Choose at random hand pressure *P1* eq. (2) Length *P2* and Width *P3* eq. (4)<br>          For each drawing<br>            if (hand pressure was low)<br>              Update the drawings utilizing eq. (5)<br>              Set *LR* and *SR* to HIGH (0.6-1)<br>            Endif<br>             Elseif (*XiGR* is near to golden ratio)<br>               Consider the learned patterns, *LR* and *SR* using eq. (6)<br>              Set *LR* and *SR* to LOW (0-0.5)<br>            Endif<br>          Evaluate the values of cost<br>          Update global best and personal best<br>          Update pattern memory<br>          Store the Best Value of cost<br>        End for<br>      $t=t+1$<br>     End while<br>    End<br>    Return the best feature vector |

Table 2 describes the detailed information of the datasets used in this paper, while Table 3 shows the parameter values of the methods used in the study. Every dataset used must be balanced. In addition, the Big Covid and Moderate Covid datasets contain image data, whereas the Breast Cancer and Iris databases are numerical datasets. Before training, characteristics from the photos in the Moderate and Big Covid datasets were retrieved, including GLCM (Gray-Level Co-occurrence Matrix) and GLDM (Gray-Level Dependence Matrix).

Table 2 List of used datasets

| No. | Dataset | Instances number | Features number |
|---|---|---|---|
| 1 | Breast Cancer | 569 | 30 |
| 2 | Moderate covid | 800 | 126 |
| 3 | Big covid | 9000 | 126 |
| 4 | Iris | 150 | 4 |

Table 3 Parameters values for used methods

| Methods | Parameters Values |
|---|---|
| BCDDO algorithm | Iteration no.: 100<br>Size of population: 30<br>Up:1<br>Lb:0<br>CR = 0.1<br>LR=0.01<br>SR=0.9<br>PS=10<br>Th_val:0.5 |
| XGboost classifier | Training set: 30%<br>Testing set: 70 % |
| KNN classifier | Number of neighbors:5<br>Number of classes:2 |

Table 4 describes the evaluation metric used to evaluate the classification by using the proposed method and the definition of each of these metrics.

Table 4 Classification performance evaluation metrics [5].

| Evaluation metric | Definition |
|---|---|
| $Accuracy$ | $\dfrac{TP + TN}{TP + TN + FP + FN}$ |
| $precision$ | $\dfrac{TP}{FP + TP}$ |
| $recall$ | $\dfrac{TP}{FN + TP}$ |
| $F1$ | $2 \times \dfrac{Specificity \times Recall}{Specificity + Recall}$ |

5. **Experimental Results and Discussions**

Four medical datasets were employed in the experiment: a breast cancer dataset with 569 data samples, a chest x-ray dataset with 800 images, and a large dataset with 9000 images. The BCDDO algorithm was used in feature selection to improve the classification of the medical data sets, and after testing and training, it demonstrated a high level of classification accuracy. The dataset was split into two parts, with 30% used for testing and validation and 70% for training. According to Table 5, which uses the Xgboost classifier, the suggested technique has a high accuracy rate. Figure 4, (a) shows the breast cancer dataset confusion matrix, while the confusion matrix for the moderate covid dataset is shown in Figure 4 (b), and Figure 4 (c) illustrates the confusion matrix for the big covid dataset while Figure 4 (d) shows the iris dataset confusion matrix.

Furthermore, preprocessing was applied to the intermediate and large COVID datasets to improve the quality of the X-ray images. To address the noise, quality, and size challenges that are inherent in X-ray images, preprocessing is essential to image enhancement. The data was prepared in several ways to aid with feature extraction and boost classification accuracy. In particular, the Contrast Limited Adaptive Histogram Equalization (CLAHE) technique was used to modify image contrast based on histograms after the Gaussian blur filter was applied for denoising [12]. The improved contrast and clarity of the resulting images from this preprocessing method made it easier to extract features later on and improved classification accuracy. We used the widely used Gray-Level Co-occurrence Matrix (GLCM) approach for texture analysis to extract features. Eight statistics—Angular Second Moment (ASM), Variance, Inverse Difference Moment (IDM), Mean, Contrast, Entropy, Homogeneity (HOM), and Correlation (COR)—are used by GLCM to gather extensive data about image textures. Pixel distance and angle functions are represented by a matrix formed by the relationship between two picture pixels and their grayscale values at different angles. This relationship is the basis for these statistics. In addition, the Gray-Level Dependence Matrix (GLDM) method was applied, which takes into account pixels with different displacements and gray-level disparities.

*Table 5 Performance of BCDDO over XGboost classifier*

| Dataset | Accuracy | Precision | Recall | F 1 | Support |
|---|---|---|---|---|---|
| **Breast Cancer** | 0.9883 | 0.990 | 0.990 | 0.990 | 171 |
| **Moderate covid** | 0.9875 | 0.990 | 0.990 | 0.990 | 240 |
| **Big covid** | 0.9996 | 1.000 | 1.000 | 1.000 | 2862 |
| **Iris** | 0.9600 | 0.959 | 0959 | 0.959 | 45 |

The above Table noted that the proposed method when used for feature selection outperforms in terms of classification with using of Four datasets. Evaluation of the BCDDO over the XGboost classifier is shown in Figure.

*Table 6 Performance of BCDDO over XGboost classifier for each class*

| Dataset | Classes Accuracy | | |
|---|---|---|---|
| | Class 0 | Class 1 | Class 2 |
| **Breast Cancer** | 0.99 | 0.98 | |
| **Moderate covid** | 0.99 | 0.98 | |
| **Big covid** | 0.99 | 0.99 | |
| **Iris** | 0.99 | 0.98 | 0.88 |

The classification accuracy for each dataset across various classifications is shown in Table 6. Across all datasets, it consistently shows high accuracy rates for Class 0 and Class 1, with near-perfect accuracy achieved by the Big COVID, Moderate COVID, and Breast Cancer datasets. Class 2 shows a little worse accuracy than the other classes for the Iris dataset.

| | |
|---|---|
| 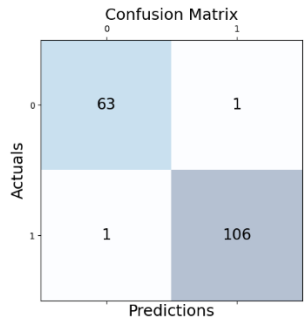 Figure 4 (a): Breast cancer dataset confusion matrix | 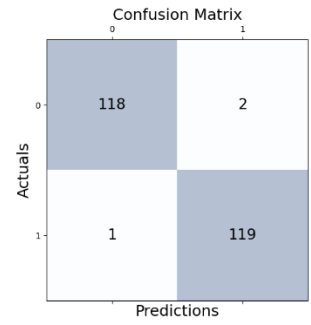 Figure 4 (b): Moderate COVID dataset confusion matrix |
| 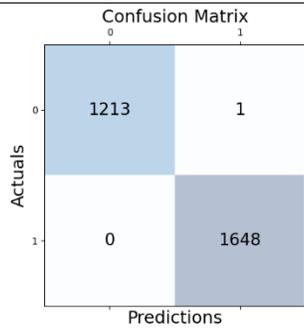 Figure 4 (c): Big Covid dataset confusion matrix | 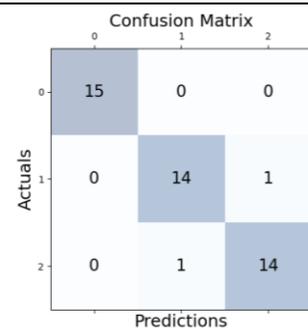 Figure 4 (d): Iris dataset confusion matrix |

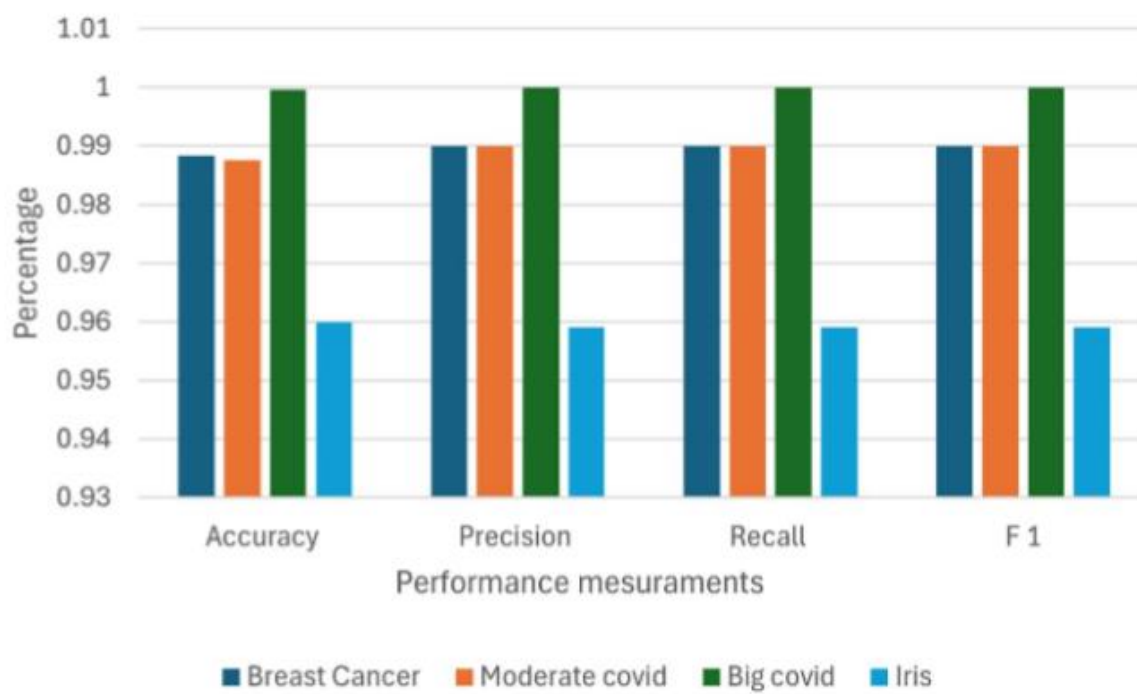

*Figure 5: Evaluation of the BCDDO over the XGboost classifier*

## 6. Comparative Study

In this portion of the experiment, the performance of the proposed BCDDO is compared further to that of HHO, SSA, Grey Wolf Optimization (GWO), Whale Optimization Algorithm (WOA), Bat algorithm (BAT), and Hybrid of Harris hawk and salp optimization (HHOSSA) algorithm. Using four datasets, the accuracy of these methods was determined. Table 7 and Figure 6, provide a detailed comparison of proposed BCDDO with HHO, SSA, GWO, WOA, BAT, and HHOSSA. As can seen BCDDO outperforms other competitor algorithms.

*Table 7: Comparison of proposed BCDDO with HHO, SSA, GWO, WOA, BAT, HHOSSA*

| Datasets | BCDDO | WHO | SSA | GWO | WOA | BAT | HHOSSA |
|---|---|---|---|---|---|---|---|
| **Breast Cancer** | **0.9883** | 0.9415 | 0.9473 | 0.9590 | 0.9422 | 0.96.8 | 0.9970 |
| **Moderate covid** | **0.9875** | 0.9464 | 0.9464 | 0.9385 | 0.96.42 | 0.97.3 | 0.9800 |
| **Big covid** | **0.9996** | 0.9213 | 0.9130 | 0.9425 | 0.9572 | 0.96.8 | 0.9880 |
| **Iris** | **0.9600** | 0.9110 | 0.9550 | 0.9460 | 0.9210 | 0.95.5 | 0.9520 |

Observing Table 7, we discover that the suggested method has obtained a significant classification accuracy advantage over a set of previous algorithms applied in the field of feature selection to improve classification accuracy.

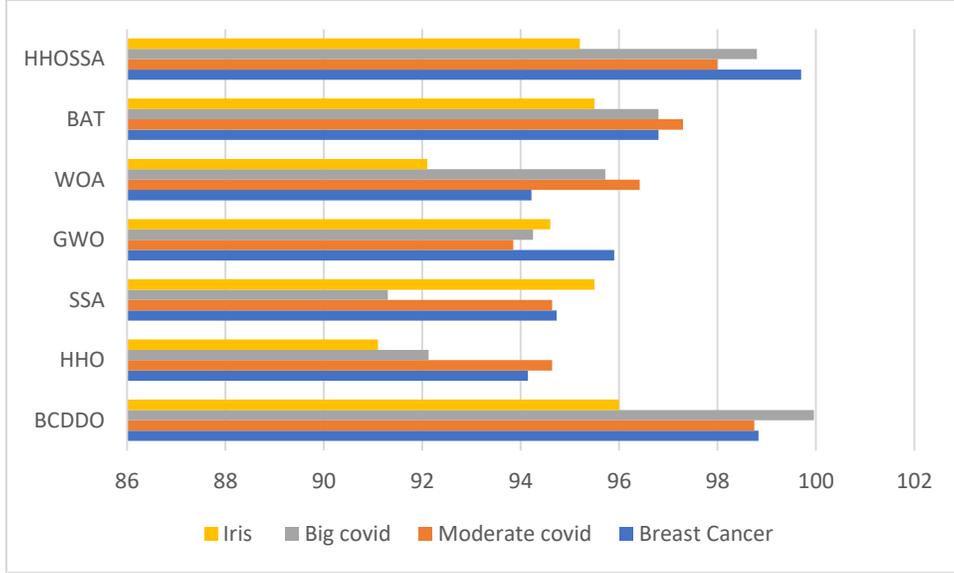

*Figure 6: Comparison of classification performances between BCDDO and art-of-states algorithms*

*Table 8: Processing time of BAT, WOA, GWO, SSA, HHO, HHOSSA, BCDDO*

| Algorithm | Average processing time (seconds) |
|---|---|
| BAT | 0.9 |
| WOA | 0.8 |
| GWO | 0.98 |
| SSA | 1.14 |
| HHO | 0.99 |
| HHOSSA | 1.06 |
| **BCDDO** | **0.7** |

The average processing times in seconds for several optimization methods, such as BAT, WOA, GWO, SSA, HHO, HHOSSA, and BCDDO, are shown in Table 8. The findings show that among the algorithms, BCDDO has the fastest processing time (0.7 seconds), closely followed by WOA (0.8 seconds), while SSA (1.14) has the longest processing time. The results reveal that BCDDO outperforms other optimization algorithms in terms of processing time, with the shortest average time of 0.7 seconds. This suggests that BCDDO is the most efficient algorithm among those compared, highlighting its superiority in terms of computational efficiency.

## 7. Conclusion

In this paper, BCDDO is proposed for FS in classification. For evaluation purposes, 3 medical datasets were used. The reported results showed that the BCDDO method has shown superior results in comparison to other similar methods by classification accuracy, also using the XGboost classifier leads to high classification precision. The efficacy of BCDDO in classification tasks and other binary optimization tasks is worth exploring in future research. Moreover, the implementation of other classification methods such as Deep Neural Decision Trees and Confidence Weighted Linear Classification using BCDDO for feature selection can also be investigated. Additionally, the study suggests applying BCDDO in other domains, such as networking and task scheduling. Also, other future work might be considered to transform other metaheuristic algorithms such as FOX [14], FDO [15], LPB [16], Leo [17], DSO [18], Goose [19], ANA [20] into binary with different applications.

**Acknowledgments:**

The authors would like to thank the University of Technology, Baghdad, and the University of Kurdistan-Hewler for providing facilities for this research work.

**Funding:** This study was not funded.

**Compliance with Ethical Standards Conflict of Interest:** The authors declare that they have no conflict of interest.

**Ethical Approval:** This article does not contain any studies with human participants or animals performed by any of the authors.

**Data Availability Statement:** The datasets generated during and/or analyzed during the current study are not publicly available but are available from the corresponding author on reasonable request